\documentclass[runningheads]{llncs}

\usepackage{eccv}

\usepackage{eccvabbrv}

\usepackage{graphicx}
\usepackage{booktabs}

\usepackage[accsupp]{axessibility}  
\usepackage[dvipsnames]{xcolor}
\usepackage{float}
\usepackage{caption}
\usepackage{geometry}
\usepackage{array}
\newcolumntype{C}[1]{p{#1}}
\usepackage{icomma}

\usepackage{hyperref}

\usepackage{multirow}
\usepackage{pifont}

\usepackage{dblfloatfix}
\usepackage{float}

\usepackage{orcidlink}

\footnotetext{Authors contributed equally.}

\DeclareRobustCommand{\myparagraph}[1]
{\noindent\textbf{#1}} 

\DeclareRobustCommand{\myparagraphit}[1]
{\noindent\textit{#1}}

\begin{document}

\title{Length-Aware Motion Synthesis\\ via Latent Diffusion} 


\author{Alessio Sampieri$^*$\orcidlink{0000-0002-1432-7499} \and
Alessio Palma$^*$\orcidlink{0009-0008-4332-9179} \and
Indro Spinelli\orcidlink{0000-0003-1963-3548}  \and
Fabio Galasso\orcidlink{0000-0003-1875-7813}}

\authorrunning{A.~Sampieri et al.}

\institute{Sapienza University of Rome, Italy \\
\email{surname@di.uniroma1.it}
}

\maketitle
\begin{abstract}
The target duration of a synthesized human motion is a critical attribute that requires modeling control over the motion dynamics and style. Speeding up an action performance is not merely fast-forwarding it. However, state-of-the-art techniques for human behavior synthesis have limited control over the target sequence length.

We introduce the problem of generating \emph{length-aware} 3D human motion sequences from textual descriptors, and we propose a novel model to synthesize motions of variable target lengths, which we dub ``Length-Aware Latent Diffusion” (\textit{LADiff}). \textit{LADiff} consists of two new modules: 1) a length-aware variational auto-encoder to learn motion representations with length-dependent latent codes; 2) a length-conforming latent diffusion model to generate motions with a richness of details that increases with the required target sequence length. \textit{LADiff} significantly improves over the state-of-the-art across most of the existing motion synthesis metrics on the two established benchmarks of HumanML3D and KIT-ML.

The code is available at \url{https://github.com/AlessioSam/LADiff}.

\end{abstract}    
\section{Introduction}\label{sec:intro}

\begin{figure}[!t]
    \centering
    \includegraphics[width=1\textwidth]{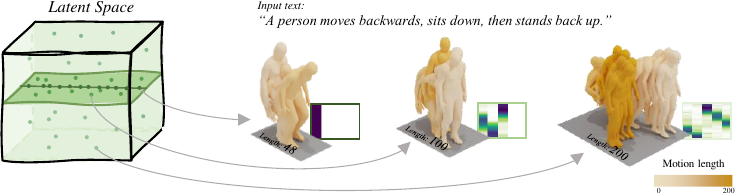}
    \caption{
    A pictorial illustration of the proposed LADiff generative model. The green latent space, learned via VAE, is subdivided into subspaces that activate progressively for longer target human motion sequences, i.e., the shortest sequence latent vectors lie on the 1D line, those for longer sequences lie on planes, cubes, or higher dimensions.
    Correspondingly, during the sequence generation via a latent DDPM, longer sequences learn attention patterns made up of more subspaces, i.e., more columns in the rectangles of latent-to-frame attention vectors. See Secs.~\ref{sec:intro} and \ref{sec:discussion} for more details.
    As a result shorter sequences depict faster actions, adopting the styles of motion that take the fewest frames. Longer sequences accommodate more frames with the longer version of the actions in terms of dynamics and style. We provide videos associated with the images in the paper in the additional materials.
    }
    \label{fig:teaser}
\end{figure}

Synthesizing human behavior is essential for creating immersive virtual worlds, whether designed for our leisure or to train robots to interact with real people safely and on a large scale \cite{puig2023habitat}. SOTA techniques for generating human behavior condition the generation on textual inputs. However, they have limited control over influential attributes, such as the desired sequence length. For example, suppose we want to generate a short kick motion. It is not enough to sub-sample a lengthy kick sequence. Instead, we should consider shorter duration and faster dynamics while respecting the laws of physics.

Leading approaches in 3D human motion generation cannot control the synthesized motion's length or consider the target sequence's variable length as a stylizing attribute. GPT-based approaches~\cite{Radford2018gpt, zhang2023t2mgpt, jiang2023motiongpt} are an example of the former, which predict auto-regressively. They can hardly constrain the output length, even when conditioned on the target length. Other leading techniques are based on diffusion models (DDPM)~\cite{Ho20ddpm} and latent representations~\cite{Rombach22latent}. One such approach is MLD~\cite{chen2023mld}, where both the Variational Autoencoder (VAE)~\cite{Kingma2014vae} representation is agnostic of the desired size of the output sequence and the diffusion-based synthesis does not account for mechanisms which affect the output sequence style, depending on the desired length. MotionDiffuse \cite{zhang2022motiondiffuse} is the sole to add a length control over sampling the noise distribution in the denoising stage of the diffusion process. However, it only adds ad-hoc mechanisms for ensuring smoothness across actions of multi-activity generations (e.g., ``synthesize a character which walks and sits'').

Our proposal, ``Length-Aware Latent Diffusion'' (\textit{LADiff}), introduces novel length-aware models both for the latent representation and for the diffusion-based synthesis to exploit the length of the target sequence as a given input. We argue that the VAE latent embedding space should encode longer sequences with larger capacity because the longer motions require more details to generate.
Therefore, in \textit{LADiff}, we organize the latent representation space into subspaces, which activate progressively with the increasing target sequence length. As depicted in Fig.~\ref{fig:teaser}, shortest sequences only lie in one subspace (on the \textit{1D green line}). Instead, longer sequences have higher-dimensional latents, which activate more subspaces (the \textit{green plane}, then the \textit{green cubic} latent space).

To cope with latent vectors of different sizes, we introduce a novel latent-length-adaptive diffusion model, which we use in \textit{LADiff} to synthesize the motions. At training time, the diffusion model attends to a latent vector of varying dimensionality. In Fig.~\ref{fig:teaser}, next to each generated sequence, the yellow-to-blue columns represent the attention of each latent to each frame. We use masking (blanked columns) for the unattended latents. 
During inference, longer sequences activate more subspaces, thus higher-dimensional latent each specializing in a chunk of frames of the extended sequence. See also the discussion in Sec.~\ref{sec:discussion}.

\textit{LADiff} is the first to bridge the \emph{noise gap} between VAEs, used for the latent representation learning, and DDPMs, adopted in the generation phase. VAEs learn to encode and decode clean motions in the first stage of representation learning. By contrast, DDPMs generate latent vectors via denoising, and their output may carry residual noise and a certain degree of stochasticity, which does not match the clean latent expected by the VAEs' decoder. To cope with the intrinsic DDPMs residual noise, we propose denoising VAEs~\cite{Im17dvae} to learn from perturbed input signals, enhancing the model's robustness against latent vector perturbations and boosting the final decoding performances.

We test \textit{LADiff} on two prominent and widely used datasets, HumanML3D~\cite{Guo22humanml3d} and KIT-ML~\cite{plappert16kit}, where it surpasses the performance of current leading techniques across multiple metrics.
Specifically, the model leverages a smaller subspace to generate shorter motion sequences, and despite this limitation, it achieves better or comparable scores compared to the SOTA. This success is attributed to utilizing all training sequences to learn the latent features within that subspace. On the contrary, for longer sequences, the model learns to correspond contiguous frames to specialized subspaces to synthesize longer sequences. Thanks to the additional capacity and this specialization, \textit{LADiff} generates convincing and realistic sequences.
See the in-depth analysis of Sec.~\ref{sec:discussion} and Fig.~\ref{fig:bin}.
 
We summarize our key contributions as follows:
\begin{itemize}
    \item Introduction of a VAEs that generates a length-aware latent space;
    \item Introduction of a length-aware DDPM and increased synergy between representation and synthesis with the introduction of DVAE for motion;
    \item An in-depth analysis of length-aware latents and related motion dynamics, and a  comprehensive evaluation of \textit{LADiff} on the HumanML3D and KIT-ML datasets, where it sets a new SOTA.
\end{itemize}

\section{Related Work}
In recent years, text-conditioned human motion synthesis has emerged as a prominent area of research. Learning a latent representation and the conditional generation are building blocks of this task, which we discuss as related work here.

\textbf{Latent representation learning.}
Sequence representation is a challenging and widespread task for representing human motion \cite{butepage2017deephuman}. This encoding can occur in either a discrete or continuous space.\\
Regarding \textit{discrete} representation, T2M-GPT~\cite{zhang2023t2mgpt} employs Vector Quantized Variational Autoencoders (VQ-VAE)~\cite{denoord17vqvae} to build a motion ``vocabulary'' with codebooks for motion generation. MotionGPT~\cite{jiang2023motiongpt} exploits VQ-VAE, combining text and motion vocabularies,  jointly learning language and motion before generating sequences. Similarly, AttT2M~\cite{Zhong23attt2m} utilizes word-level and sentence-level features to learn the cross-modal relationship between text and motion. M2DM ~\cite{Kong23m2dm} applies an orthogonal regularization that encourages diversity and increased usage of codebook entries to improve performance.\\
Considering \textit{continuous} latent representation, Variational Autoencoders (VAE)~\cite{Kingma2014vae} are the SOTA due to their ease of training \cite{guo2020a2m, petrovich21actor}. TEMOS \cite{petrovich22temos} introduces a Transformer-based VAE model \cite{vaswani2017transformer} that leverages a sequence-level latent vector to generate text-conditioned motions. MLD \cite{chen2023mld} employs a VAE to learn latent representations of motion sequences. HumanML3D \cite{Guo22humanml3d} uses a temporal VAE to enhance the quality of fixed-size latent representations and incorporates an MLP for predicting sequence length from textual descriptions.

Our approach builds upon the success of VAE for its representation performance and ease of training. However, we organize its latent space in different subspaces to encode sequences of different lengths. No prior work, neither continuous nor discrete, designs latent representations that are aware of target lengths.

\textbf{Generation.} In this fast-evolving field, first approaches such as TEMOS~\cite{petrovich22temos} directly used VAEs to generate motion based on textual conditioning, where the number of positional encodings established during generation determines the duration of these motions. There are two leading generation strategies: Generative Pre-trained Transformers (GPT) \cite{Radford2018gpt}, and Denoising Diffusion Probabilistic Models (DDPM)~\cite{Ho20ddpm}. The first, represented by approaches such as MotionGPT ~\cite{zhang2023t2mgpt} and T2M-GPT\cite{jiang2023motiongpt} leverages GPT architectures for autoregressive generation, where the occurrence of a specific token marks the generation's end. 
MDM~\cite{tevet2023mdm} is the archetype of the latter family of models, where that apply DDPM directly to raw motion data to capture the motion-text relationship for a determined number of tokens. MotionDiffuse~\cite{zhang2022motiondiffuse} adds smoothing to body parts and frames during noise sampling, which can be resource-intensive. 
Leveraging latent DDPM~\cite{Rombach22latent}, MLD \cite{chen2023mld} achieves SOTA performances. It presents a two-step approach wherein they initially use VAEs to learn a compact and low-dimensional motion latent space, followed by a diffusion process in this space.
M2DM~\cite{Kong23m2dm} mixes GPT and DDPM approaches by using the latter to synthesize motions starting from a discrete latent space built using a token priority mechanism and a transformer-based decoder.
We build upon the success of MLD \cite{chen2023mld}. Still, we organize the latent space into subspaces dedicated to motions of different lengths, adapting the conditional DDPM generation to the variability of the target length to have a time-dependent style and dynamics.\\
Retrieveal Augmented Generation~\cite{lewis20rag,zhang2023remodiffuse} has shown significant performance improvements when conditioning motion generation on a sample retrieved from the entire training set. However, this approach requires storing and searching through a large database. We demonstrate how the length-aware generation of \textit{LADiff} can also be conditioned on the same retrieved sample.

\section{Methodology}
\label{sec:methodology}

In this Section, we introduce the length-aware VAE for learning length-aware latent representations and latent diffusion process. Also, here we detail the proposed Denoising VAE to add robustness during synthesis. Next, we describe background concepts on the latent space, generation, and conditioning.

\begin{figure}[!h]
    \centering
    \includegraphics[width=1\textwidth]{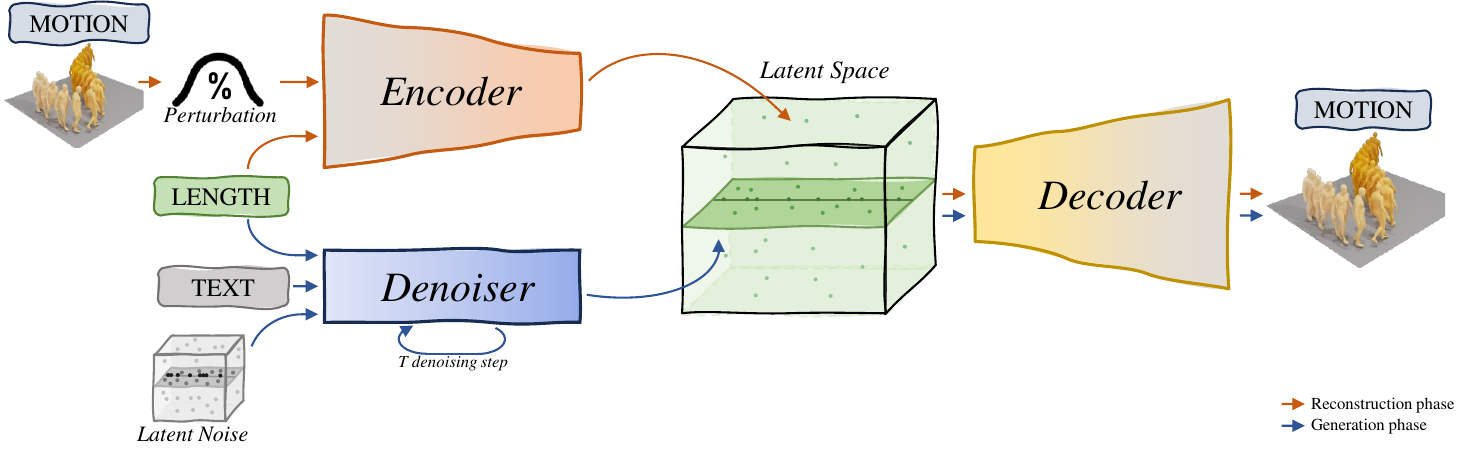}
    \caption{Overview of our proposed Length-Aware Latent Diffusion (\textit{LADiff}). During the reconstruction phase (orange arrows), the Encoder, aided by the Decoder, learns to represent sequences of varying lengths into a latent space composed of subspaces, which activate progressively for longer sequences. In the Generation stage (blue arrows), the Denoiser learns to create latent vectors aligned to the textual input, which map to correct subspaces specified by input sequence length. The actual motion results from decoding the latent vectors of the Denoiser. For this purpose, the Decoder is made resilient to noise in the Reconstruction stage by learning to reconstruct sequences affected by noise}
    \label{fig:pipeline}
\end{figure}

\paragraph{Problem formalization.} The objective of motion synthesis is to generate a motion $\mathbf{X}$ based on textual input $\mathbf{w}$. We define motion as a sequence of poses $\mathbf{X}=[X_1, .., X_F ] \in \mathbb{R}^{F\times V}$, where $F$ are the number of frames and $V$ the parameters of the pose vector. 
Following literature~\cite{chen2023mld, zhang2023t2mgpt, jiang2023motiongpt, Guo22humanml3d}, our pose vector $V$ contains pose, velocity and rotation. We also define the textual description as a vector $\mathbf{w}\in \mathbb{R}^{1 \times D}$. In this work, we also provide as input the target sequence length $ f^* \in \mathbb{R}$.

\subsection{Background}
Our approach is based on the current best practice in motion synthesis, where motions are generated using DDPM~\cite{tevet2023mdm, Kong23m2dm, chen2023mld, zhang2022motiondiffuse}, often with VAEs trained for reconstruction tasks. In this context, textual inputs condition the synthesis. First, the model reconstructs the input motion sequence, crafting a meaningful latent space. Then, the model generates the output motion guided by textual input.

\paragraph{Latent representation learning.} VAE approaches consist of an encoder-decoder generative architecture learned to minimize a reconstruction error. In this framework, latent representations $\mathbf{Z}$ are the lower-dimensional embeddings produced as output by the encoder network parametrized by $\phi$ from the input data $\mathbf{X}$. Following VAE literature \cite{Kingma2014vae}, $q_\phi(\mathbf{z} \vert \mathbf{x})$ approximates the true posterior distribution of the latent space with multivariate Gaussian with a diagonal covariance structure:
\begin{equation}
    q_\phi(\mathbf{z} \vert \mathbf{x}) = \mathcal{N}(\mathbf{z} \vert \mu_\phi(\mathbf{x}), \sigma^2_\phi(\mathbf{x})\mathbf{I}),
\end{equation}
where $\mu_\phi(\cdot)$ and $\sigma^2_\phi(\cdot)$, are outputs of the encoder. We sample from the approximate posterior $\mathbf{z}^{(i)} \sim q_\phi(\mathbf{z}\vert \mathbf{x}^{(i)})$ using: 
\begin{equation}
\label{eq:sampling}
    \mathbf{z}^{(i)} = \boldsymbol{\mu}^{(i)} + \boldsymbol{\sigma}^{(i)2} \odot \boldsymbol{\rho},
\end{equation}
where $\mathbf{z}^{(i)} \in \mathbb{R}^{1 \times D}$ is the latent vector, $\boldsymbol{\rho} \sim \mathcal{N}(0,\mathbf{I})$ is additional noise, $\odot$ represent pairwise multiplication and $\mathbf{x}^{(i)}$ is a sample from the dataset $\mathbf{X}$.
The decoder $p_\theta( \mathbf{x} \vert \mathbf{z})$ is a network parametrized by $\theta$ that maps the sampled values back to the input space. The parameters of the networks are obtained by optimizing the ELBO objective as described in \cite{Kingma2014vae}.

\paragraph{Generation.} Latent DDPM have a different approximated posterior $g(\mathbf{z}_t \vert \mathbf{z}_{t-1})$, denoted \textit{diffusion process}, which gradually gradually converts latent representations $\mathbf{z}_{0}=\mathbf{z}$ into random noise $\mathbf{z}_{T}$ in $T$ timesteps: 

\begin{equation}
    g(\mathbf{z}_t \vert \mathbf{z}_{t-1}) = \mathcal{N}(\mathbf{z}_t; \sqrt{\bar{\alpha}_t}\mathbf{z}_{t-1}, (1-\bar{\alpha}_t)\mathbf{I})\,,
    \label{eq:diffusion}
\end{equation}
where $\bar{\alpha}_t$ is a scaling factor specific to timestep $t$.
Then, the reverse process dubbed \textit{denoising} gradually refines the noised vector to a suitable latent representation $\mathbf{z}_{0}$. Following \cite{chen2023mld, Rombach22latent, Ho20ddpm, dhariwal21, Saharia23}, we use the notation $\{\mathbf{z}_t\}^T_{t=0}$ to denote the sequence of noised latent vectors, with $\mathbf{z}_{t-1} = \epsilon_\psi(\mathbf{z}_t, t)$ representing the denoising operation at time step $t$ with $\epsilon_\psi$ being a denoising autoencoder trained to predict the denoised variant of its input.

\paragraph{Conditioning.} Following literature~\cite{chen2023mld, Gu22vqdiff,Balaji2022eDiffITD}, we incorporate textual input for conditional synthesis, enhancing the overall control and expressiveness of the generated motion sequences. The objective is to align textual and motion representations:
\begin{equation}
    \mathcal{L} = \mathbb{E}_{\epsilon \sim \mathcal{N}(0,\mathbf{I}),t, \mathbf{w}}[ \| \epsilon - \epsilon_{\psi} (\mathbf{z}_t,t,\gamma(\mathbf{w})) \|^2_2 ]\,,
    \label{eq:diffusion_opt}
\end{equation}
where we add the textual input encoded via \textit{CLIP-ViT-L-14} \cite{ramesh22clip} $\gamma(\mathbf{w})$.

\subsection{Length-Aware VAE}
We propose a novel Length-Aware VAE which extends the VAE framework by organizing the latent space to characterize the motion's length constraint.
We decompose the entire latent space into $K$ subspaces. Therefore, the complete latent space has dimension $\mathbb{R}^{K\times D}$, and the smallest dimensional subspace is $1 \times D$-dimensional. On the latter live the latent vectors associated with the shortest sequences. As the motion length grows, we stepwise unlock bigger subspaces following the activation rate $k=\lceil \frac{f}{r} \rceil$, where $k$ is the number of activated subspaces, $f$ represents the number of frames, and $r$ is the number of frames assigned to each subspace. This results in latent vector dimensionality that grows with the sequence length $f$, exploiting $k$ subspaces of the latent space, each encoding exactly $r$ frames.
In Fig. \ref{fig:teaser}, we visualize the shortest motions on a 1-dimensional subspace represented by a line. The 2-dimensional plane represents middle-sized sequences and comprises the previous subspace. Finally, the most extended sequences exploit the entire 3-dimensional space enclosing the previous subspaces.
We train the encoder-decoder Length-Aware VAE by passing the motion sequence $\mathbf{X}$ through the encoder, obtaining the desired $k$ means ($\boldsymbol{\mu}$) and variances ($\boldsymbol{\sigma}^2$):
\begin{equation}
    q_\phi(\mathbf{z} \vert \mathbf{x}) = \mathcal{N}(\mathbf z \vert \mu_1(\mathbf{x}), \sigma^2_1(\mathbf{x})\mathbf{I},\ldots, \mu_k(\mathbf{x}), \sigma^2_k(\mathbf{x})\mathbf{I})
\end{equation}

Then we sample $k$ values according to the activated subspaces to build our latent representation $\mathbf{z} \in \mathbb{R}^{k \times D}$ though concatenation:
\begin{equation}
    \mathbf{z}^{(i)} = [\boldsymbol{\mu}^{(i)}_1 + \boldsymbol{\sigma}^{(i)2}_1 \odot \boldsymbol{\rho}_1,\ldots, \boldsymbol{\mu}^{(i)}_k + \boldsymbol{\sigma}^{(i)2}_k \odot \boldsymbol{\rho}_k] \,.
\end{equation}
Finally, we use the decoder to reconstruct the input sequence $\mathbf{X}$. The transformer-based decoder handles the varying dimensional space by using the attention mechanism. The training of the encoder and decoder represents the first stage of our training pipeline.
From these procedures arises a structured latent space with subspaces specialized on sequences of different lengths. The first subspace, dedicated to the shortest sequences, has the smallest dimensional space. However, it is active and trained for every sequence in the dataset. The last subspaces are engaged only for longer sequences with two main consequences: first, longer sequences have a higher capacity space, and second, they specialize over a subset of the entire data distribution containing longer sequences.

\subsection{Length-Aware Latent Diffusion}

We present the core concept of our Length-Aware Latent Diffusion process: the dynamic adaptation of latent dimensions for generating motion sequences.
Specifically, each latent $\mathbf{z}_t$ resides in $\mathbb{R}^{k \times D}$, with $k \in \{1,\ldots,K\}$. During sampling (or at inference time), we initialize $\mathbf{z}_T$ using our activation rate $k$ and the desired motion length $f^*$:
\begin{equation}
    \mathbf{z}_T = \mathcal{N(\mathbf 0, \mathbf I)} \in \mathbb{R}^{\lceil \frac{f^*}{r} \rceil \times D}
\end{equation}

The VAE's decoder receives the denoised vector $\mathbf{z}_{\mathbf{0}} \in \mathbb{R} ^{k \times D}$ together with the desired motion length $f^*$ for the final synthesis. The transformer decoder uses attention to account for the generated latents of variable length. As a byproduct of our adaptive latent vectors, the inference time is reduced, since the decoder uses fewer terms in the attention mechanisms.
The length-aware latent diffusion denoiser is trained in a second stage, during which the encoder and decoder are kept frozen.

\subsection{Denoising VAE}
We train VAE and latent-DDPM in cascaded stages. However, they assume distinct characteristics of the latent vectors. We train VAEs to reconstruct clean inputs as faithfully as possible. Conversely, the latent DDPMs generate latent vectors from pure Gaussian noise. These generated latent vectors may contain residual noise, which poses challenges for the VAE decoder, trained to reconstruct clean inputs and then frozen.

To establish a more coherent connection between these two phases, we propose employ Denoising VAE (DVAE)~\cite{Im17dvae} to perturb features of the input sequence to create a robust latent space. We randomly sample a percentage of frames from the input sequence and perturb them with controlled Gaussian noise $\epsilon\sim\mathcal{N}(0,1)$. Therefore, we introduce an augmentation strategy to align the latent variable distribution from the two phases.

\subsection{Implementation Details} 
\label{implementation_details}
The framework comprises three major blocks: the encoder $\mathcal{E}$, decoder $\mathcal{D}$  and denoiser $\epsilon_\theta$. While  $\mathcal{E}$ and $\mathcal{D}$ constitute the reconstruction phase (24M parameters), the denoiser (9M parameters) together with the frozen decoder $\mathcal{D}$ form the generation phase. Each block consists of 9 layers and 4 heads, with an embedding dimension of $D=256$. Both the encoder and decoder are constructed using standard transformer layers. The denoiser $\epsilon_\theta$ incorporates classical self-attention over the latent vector and linear cross-attention between latent vector and textual input, adopting also the Stylization block from~\cite{zhang2022motiondiffuse}. 
We use a batch size of 64 for the first phase and 128 for the second. The learning rate is set to $10^{-4}$, and the optimizer used is AdamW~\cite{loshchilov2018adamw}. During training, we performed 1000 diffusion steps. At inference, we use only 20 diffusion steps. We used 8 Tesla V100 GPUs, conducting 3k epochs for both phases.

\section{Results}

We evaluate the reconstruction and generation capabilities of \textit{LADiff} quantitatively and qualitatively. The analysis comprehends an ablation study of the proposed model components.

\subsection{Datasets and Metrics}
For this task, we evaluate our model on two large and established datasets, namely HumanML3D and KIT-ML. Both datasets utilize the SMPL~\cite{Matthew15smpl} representation with 22 and 21 joints respectively.
\textbf{HumanML3D}~\cite{Guo22humanml3d} with 14.6k motion sequences, each associated with three textual descriptions, provides an extensive range of movements, from martial arts to everyday actions. \textbf{KIT-ML}~\cite{plappert16kit} instead consists of 3.9k motions with a total of 6.2k textual descriptions.  Following \cite{zhang2023t2mgpt, chen2023mld, zhang2022motiondiffuse}, we use motions of the same lengths as HumanML3D.

\begin{table}[!tp]
\caption{HumanML3D results for motion synthesis (above) and RAG (below).}
\resizebox{1.\textwidth}{!}{
\begin{tabular}{lccccccc}
\hline
\multicolumn{1}{c}{}                          & \multicolumn{3}{c}{R Precision $\uparrow$}                                                                                                &                                 &                                          &                                 &                                  \\ \cline{2-4}
\multicolumn{1}{c}{\multirow{-2}{*}{Methods}} & Top 1                                    & Top 2                                    & Top 3                                    & \multirow{-2}{*}{FID $\downarrow$}           & \multirow{-2}{*}{MM-Dist $\downarrow$}                     & \multirow{-2}{*}{Diversity $\rightarrow$}     & \multirow{-2}{*}{MModality $\uparrow$}      \\ \hline
{Real}                   & {0.511}$^{\pm.003}$             & {0.703}$^{\pm.003}$             & {0.797}$^{\pm.002}$             & {0.002}$^{\pm.000}$    & 2.974$^{\pm.008}$             & {9.503}$^{\pm.065}$    & {-}            \\ \hline 
 \hline
MotionDiffuse~\cite{zhang2022motiondiffuse}                          & 0.491$^{\pm.001}$                                    & {0.681}$^{\pm.001}$                                    & {0.782}$^{\pm.001}$                                    & 0.630$^{\pm.001}$                            & 3.113$^{\pm.001}$                                    & 9.410$^{\pm.049}$                        & 1.553$^{\pm.042}$                            \\
MDM~\cite{tevet2023mdm}                                           & 0.320$^{\pm.005}$                                    & 0.498$^{\pm.004}$                                    & 0.611$^{\pm.007}$                                    & 0.544$^{\pm.044}$                           & 5.566$^{\pm.027}$                                    & \underline{9.559}$^{\pm.086}$                           & \underline{2.799}$^{\pm.072}$                            \\
MLD~\cite{chen2023mld}                                          & 0.481$^{\pm.003}$                                    & 0.673$^{\pm.003}$                                    & 0.772$^{\pm.002}$                                    & 0.473$^{\pm.013}$                           & 3.196$^{\pm.010}$                                    & 9.724$^{\pm.082}$                           & {2.413}$^{\pm.079}$                            \\
T2M-GPT~\cite{zhang2023t2mgpt}                                          & 0.491$^{\pm.003}$                                    & 0.680$^{\pm.003}$                                     & 0.775$^{\pm.002}$                                    & {0.116}$^{\pm.004}$                           & 3.118$^{\pm.011}$                                    & 9.761$^{\pm.081}$                           & 1.856$^{\pm.011}$                            \\
MotionGPT~\cite{jiang2023motiongpt}                                    & {0.492}$^{\pm.003}$                                    & {0.681}$^{\pm.003}$                                    & 0.778$^{\pm.002}$                                    & 0.232$^{\pm.008}$                           & {3.096}$^{\pm.008}$                                    & \textbf{9.528}$^{\pm.071}$                           & 2.008$^{\pm.084}$                            \\ 

Fg-T2M~\cite{Wang23fgt2m}                                    & {0.492}$^{\pm.002}$                                    & {0.683}$^{\pm.003}$                                    & 0.783$^{\pm.002}$                                    & 0.243$^{\pm.019}$                           & {3.109}$^{\pm.007}$                                    & {9.278}$^{\pm.072}$                           & 1.614$^{\pm.049}$                            \\ 

M2DM~\cite{Kong23m2dm}                                    & {0.497}$^{\pm.003}$                                    & {0.682}$^{\pm.002}$                                    & 0.763$^{\pm.003}$                                    & 0.352$^{\pm.005}$                           & {3.134}$^{\pm.010}$                                    & {9.926}$^{\pm.073}$                           & \textbf{3.587}$^{\pm.072}$                            \\ 

AttT2M~\cite{Zhong23attt2m}             & \underline{0.499}$^{\pm.006}$                                    & \underline{0.690}$^{\pm.002}$                                    & \underline{0.786}$^{\pm.002}$                                    & \underline{0.112}$^{\pm.006}$                           & \textbf{3.038}$^{\pm.007}$                                    & {9.700}$^{\pm.090}$                           & 2.452$^{\pm.051}$                            \\ 
Ours$_{\text{(\textit{r}=32)}}$                                     & \multicolumn{1}{c}{{0.493$^{\pm.002}$}} & \multicolumn{1}{c}{{0.686}$^{\pm.002}$} & {0.784$^{\pm.001}$}       & \textbf{0.110$^{\pm.004}$}       & {3.077$^{\pm.010}$}       & 9.622$^{\pm.071}$                & 2.095$^{\pm.076}$ \\

Ours$_{\text{(\textit{r}=48)}}$                                      & \multicolumn{1}{c}{\textbf{0.503$^{\pm.002}$}} & \multicolumn{1}{c}{\textbf{0.696}$^{\pm.003}$} & \multicolumn{1}{c}{\textbf{0.792$^{\pm.002}$}} & \multicolumn{1}{c}{{0.182}$^{\pm.004}$} & \multicolumn{1}{c}{\underline{3.054}$^{\pm.008}$} & \multicolumn{1}{c}{9.795$^{\pm.076}$} & \multicolumn{1}{c}{2.115$^{\pm.063}$} \\ \hline  \hline
ReMoDiffuse~\cite{zhang2023remodiffuse}                          & \textbf{0.510}$^{\pm.005}$                                    & \textbf{0.698}$^{\pm.006}$                                    & \textbf{0.795}$^{\pm.004}$                                    & 0.103$^{\pm.004}$                            & \textbf{2.974}$^{\pm.016}$                                    & 9.081$^{\pm.075}$                        & 1.795$^{\pm.028}$                            \\ 
Ours$_{\text{(\textit{r}=48)}}$  RAG                       &             0.494$^{\pm.002}$                      &  0.691$^{\pm.003}$   & {0.786}$^{\pm.001}$                                    & \textbf{0.054}$^{\pm.002}$                            & 3.112$^{\pm.008}$                                    & \textbf{9.517}$^{\pm.077}$                        & \textbf{2.453}$^{\pm.074}$                            \\ \hline
\end{tabular}}
\label{tab:hml3d}
\end{table}

\myparagraph{Metrics.} We evaluate the generated motion sequences employing all commonly adopted metrics. 

\myparagraphit{R-precision and MM-Dist.} In the context of the text-to-motion task, \cite{Guo22humanml3d} offers motion and text feature extractors designed to generate geometrically consistent features for paired text-motion combinations and vice versa. In this feature space, we assess motion-retrieval precision (R-precision) by mixing the generated motion with 31 mismatched motions. We then calculate the text-motion top-1/2/3 matching accuracy. Additionally, we measure Multi-modal Distance (MM-Dist), which quantifies the distance between the generated motions and the associated text. The L2-loss is employed to determine both R-Precision and MM-Dist.

\myparagraphit{FID.} We assess Motion Quality using the Frechet Inception Distance (FID) metric, which measures the similarity between the distributions of generated and real motions. It is computed by measuring the L2-loss of the latent representations obtained through the feature extractor.

\myparagraphit{Diversity and MModality.} We utilize Diversity and MultiModality (MModality) metrics to gauge the range of motion variations across the entire dataset and the diversity of generated motions for each specific text input.
To evaluate diversity, we randomly divide the data into $\{x_1, \dots, x_{X_d}\}$ and $\{x'_1, \dots, x'_{X_d}\}$, two equal-sized subsets each containing motion feature vectors.\\
To assess MultiModality, we randomly select a set of text descriptions from the entire collection, with a total of $T_d$ descriptions. Following this, we randomly sample two equal-sized subsets, $X_d$, from all the motions generated by the t-th text descriptions. These subsets contain motion feature vectors: $\{x_{t,1}, \dots, x_{t,X_d}\}$ and $\{x'_{t,1}, \dots, x'_{t,X_d}\}$, respectively. Diversity and MModality are then defined as the average L2-loss between corresponding samples.

\begin{table}[!tp]
\caption{KIT-ML results for motion synthesis (above) and RAG (below).}
\resizebox{1.\textwidth}{!}{
\begin{tabular}{lccccccc}
\hline
\multicolumn{1}{c}{}                          & \multicolumn{3}{c}{R Precision $\uparrow$}                                                                                                &                                 &                                          &                                 &                                  \\ \cline{2-4}
\multicolumn{1}{c}{\multirow{-2}{*}{Methods}} & Top 1                                    & Top 2                                    & Top 3                                    & \multirow{-2}{*}{FID $\downarrow$}           & \multirow{-2}{*}{MM-Dist $\downarrow$}                     & \multirow{-2}{*}{Diversity $\rightarrow$}     & \multirow{-2}{*}{MModality $\uparrow$}      \\ \hline
{\color[HTML]{333333} Real}                   & {\color[HTML]{333333} 0.424}$^{\pm.005}$ & {\color[HTML]{333333} 0.649}$^{\pm.006}$ & {\color[HTML]{333333} 0.779}$^{\pm.006}$ & {\color[HTML]{333333} 0.031}$^{\pm.004}$ & {\color[HTML]{333333} 2.788}$^{\pm.012}$ & {\color[HTML]{333333} 11.08}$^{\pm.097}$ & {-}        \\ \hline \hline
MotionDiffuse~\cite{zhang2022motiondiffuse}                                 & {0.417}$^{\pm.004}$                        & 0.621$^{\pm.004}$                        & 0.739$^{\pm.004}$                        & 1.954$^{\pm.062}$                        & 2.958$^{\pm.005}$                        & \textbf{11.10}$^{\pm.143}$               & 0.730$^{\pm.013}$                       \\
MDM~\cite{tevet2023mdm}                                        & 0.164$^{\pm.004}$                        & 0.291$^{\pm.004}$                        & 0.396$^{\pm.004}$                        & 0.497$^{\pm.021}$                        & 9.191$^{\pm.022}$                        & 10.85$^{\pm.109}$                        & 1.907$^{\pm.214}$                       \\
MLD~\cite{chen2023mld}                                           & 0.390$^{\pm.008}$                        & 0.609$^{\pm.008}$                        & 0.734$^{\pm.007}$                        & \textbf{0.404}$^{\pm.027}$               & 3.204$^{\pm.027}$                        & 10.80$^{\pm.117}$                        & {2.192}$^{\pm.071}$              \\
T2M-GPT~\cite{zhang2023t2mgpt}                                           & 0.416$^{\pm.006}$                        & {0.627}$^{\pm.006}$                        & {0.745}$^{\pm.006}$                        & 0.514$^{\pm.029}$                        & \underline{3.007}$^{\pm.023}$                        & {10.92}$^{\pm.108}$               & 1.570$^{\pm.039}$                       \\
MotionGPT~\cite{jiang2023motiongpt}                                     & 0.366$^{\pm.005}$                        & 0.558$^{\pm.004}$                        & 0.680$^{\pm.005}$                        & 0.510$^{\pm.016}$                        & 3.527$^{\pm.021}$                        & 10.35$^{\pm.084}$                        & \underline{2.328}$^{\pm.117}$              \\

Fg-T2M~\cite{Wang23fgt2m}                                     & \underline{0.418}$^{\pm.005}$                        & 0.626$^{\pm.006}$                        & 0.745$^{\pm.004}$                        & 0.571$^{\pm.047}$                        & 3.114$^{\pm.015}$                        & 10.93$^{\pm.083}$                        & {1.019}$^{\pm.029}$              \\

M2DM~\cite{Kong23m2dm}                                    & {0.416}$^{\pm.004}$                                    & {0.628}$^{\pm.004}$                                    & 0.743$^{\pm.004}$                                    & 0.515$^{\pm.029}$                           & {3.015}$^{\pm.017}$                                    & {11.41}$^{\pm.097}$                           & \textbf{3.325}$^{\pm.037}$                            \\ 

AttT2M~\cite{Zhong23attt2m}             & {0.413}$^{\pm.006}$                                    & \underline{0.632}$^{\pm.006}$                                    & \underline{0.751}$^{\pm.006}$                                    & {0.870}$^{\pm.039}$                           & {3.039}$^{\pm.021}$                                    & \underline{10.96}$^{\pm.123}$                           & 2.281$^{\pm.047}$                            \\ 
Ours$_{\text{(\textit{r}=48)}}$                                          & \textbf{0.429}$^{\pm.007}$       & \textbf{0.647}$^{\pm.004}$       & \textbf{0.773}$^{\pm.004}$       & \underline{0.470}$^{\pm.016}$                & \textbf{2.831}$^{\pm.020}$       & 11.30$^{\pm.108}$                & 1.243$^{\pm.057}$               \\ \hline \hline
ReMoDiffuse~\cite{zhang2023remodiffuse}                          & \textbf{0.427}$^{\pm.014}$                                    & \textbf{0.641}$^{\pm.004}$                                    & \textbf{0.765}$^{\pm.055}$                                    & \textbf{0.155}$^{\pm.006}$                            & \textbf{2.814}$^{\pm.0.12}$                                    & 10.80$^{\pm.105}$                        & 1.239$^{\pm.028}$                            \\ 
Ours$_{\text{(\textit{r}=48)}}$ RAG                        &            0.415$^{\pm.006}$                 &   0.632$^{\pm.007}$  & {0.758}$^{\pm.005}$                                    & 0.386$^{\pm.003}$                            & 2.978$^{\pm.020}$                                    & \textbf{11.20}$^{\pm.008}$                        & \textbf{1.732}$^{\pm.066}$                            \\ \hline
\end{tabular}}

\label{tab:kit}
\end{table}

\myparagraph{Baselines.} We compare our proposed \textit{LADiff} with the current state-of-the-art techniques for motion synthesis. MotionDiffuse~\cite{zhang2022motiondiffuse} and MDM~\cite{tevet2023mdm} are DDPM-based models that approach the task by directly processing raw poses. MLD~\cite{chen2023mld} is a two-stage process that uses VAE for encoding the input sequence and a latent diffusion model for conditional generation. T2M-GPT~\cite{zhang2023t2mgpt} and MotionGPT~\cite{jiang2023motiongpt} utilize GPT as the generation mechanism, leveraging VQ-VAE for motion encoding into a codebook. Fg-T2M~\cite{Wang23fgt2m} combines linguistics-structure and context-aware progressive reasoning through graph attention network~\cite{velickovic18gat}. M2DM~\cite{Kong23m2dm} encodes the sequence using a codebook and implements a generation process through a discrete diffusion model. AttT2M~\cite{Zhong23attt2m} exploits a TCN that decodes the features learned through local and global attention between the sequence and text. ReMoDiffuse~\cite{zhang2023remodiffuse} generates human motion from text using RAG~\cite{lewis20rag}, it retrieves relevant motion samples from a training set (assumed to be available at inference), and then uses a transformer to fuse them with the text input and generate motion sequences. We denote with ``Real'' the metrics obtained with the ground-truth motion.

\subsection{Quantitative Results}
We validate our model on the HumanML3D and KIT-ML datasets. Following \cite{Guo22humanml3d}, we calculate the results based on 20 generations and report the average with a 95\% confidence interval.\\

\myparagraph{Generation.} Table~\ref{tab:hml3d} shows the results obtained on the HumanML3D dataset. Our model outperforms the current state-of-the-art techniques in all R-Precision scores and FID while maintaining a comparable Matching Score. In the bottom part of the table, we report the SOTA retrieval augmented~\cite{zhang2023remodiffuse}, compare to \textit{LADiff}, augmented with retrieval. Both techniques additionally assume access to the entire training set. \textit{LADiff} enhances the FID by 47\%, exhibiting superior Diversity and MModality. 

\begin{figure}[!tp]
    \centering
    \includegraphics[width=1.\textwidth]{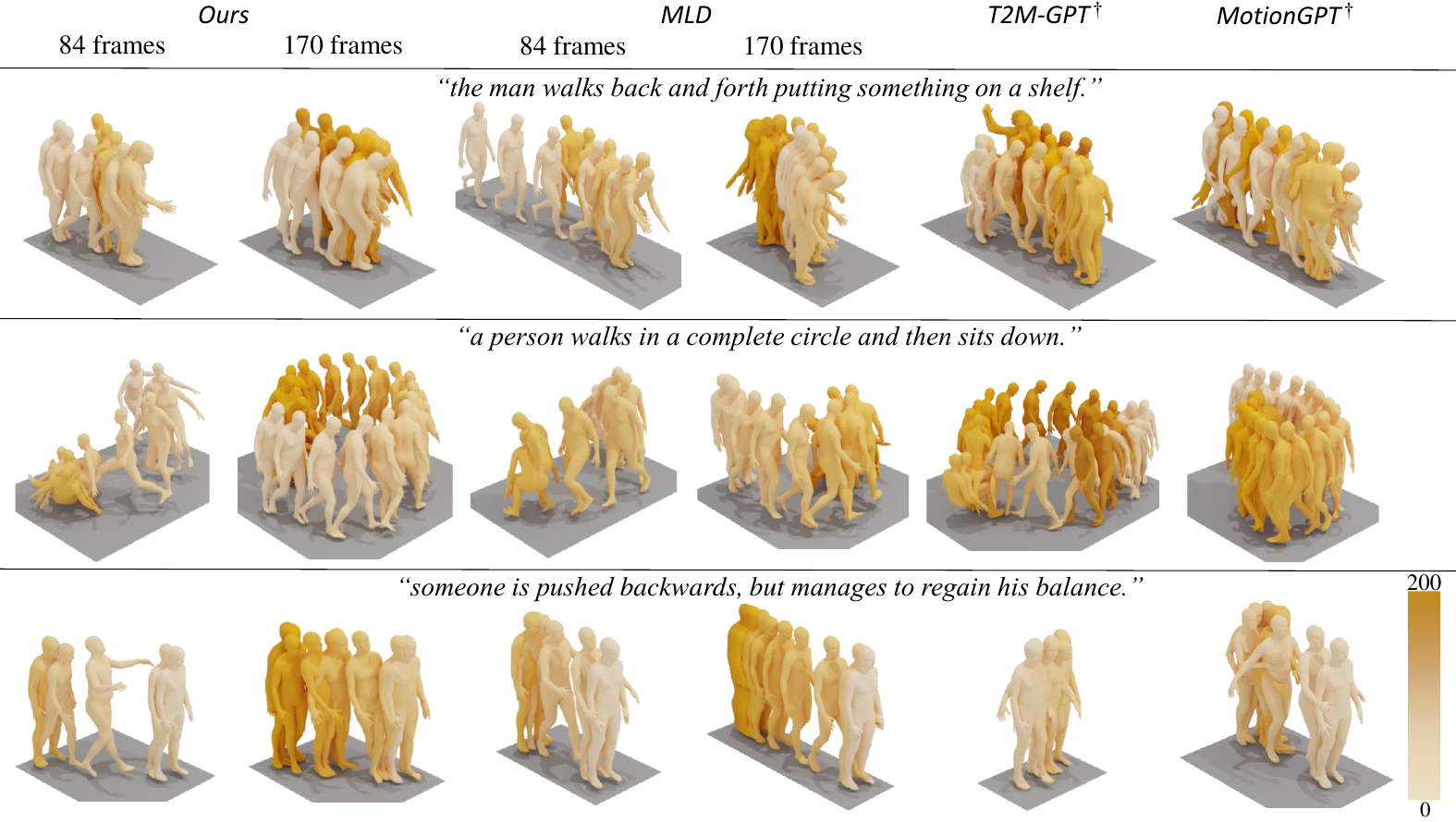}
    \caption{Qualitative comparison of text-based human motion generations. multiple target lengths are provided to techniques that allow to set the length input. See Section \ref{sec:qualitative} for discussion.}
    \label{fig:difference}
\end{figure}

The performance exceeds the top models even on the KIT-ML dataset (Table~\ref{tab:kit}). \textit{LADiff} outperform the current best techniques of 4.2\% in Matching Score and 2.2\% in R-Precision Top3. In the bottom part of the table, our method outperform \cite{zhang2023remodiffuse} in Diversity and MModality. 

\textit{LADiff} demonstrates consistent performance across different datasets in several metrics, including R precision, Matching score, and FID. These results and the R-precision, more specifically, demonstrate that \textit{LADiff} can generate novel test motions from text that are most similar to the real ones.

\subsection{Qualitative Results}
\label{sec:qualitative}
First we show qualitatively the results obtained from \textit{LADiff} and then we demonstrate how our model adapts to different lengths.\\
\myparagraph{Comparisons.} From Fig.~\ref{fig:difference} we note that \textit{LADiff} is the only framework to consider the desired length and to adapt to it, in terms of style and dynamics. GPT-based approaches have no control over the length of the output sequence. MLD can generate sequences of a desired length but lacks adaptation to the target length. See, for example, the second row, where it seems that MLD mainly fits the target length by subsampling without adapting style and dynamics. Our model is capable of rich and realistic behavior, respecting the textual conditioning and the target length (e.g., a perfect circle, realistic balance regain, correct shelf height). In the second row we see that our ``man" runs to complete the circle when the number of frames is insufficient and walks when feasible.

\begin{figure}[!ht]
    \centering
    \includegraphics[width=.95\textwidth]{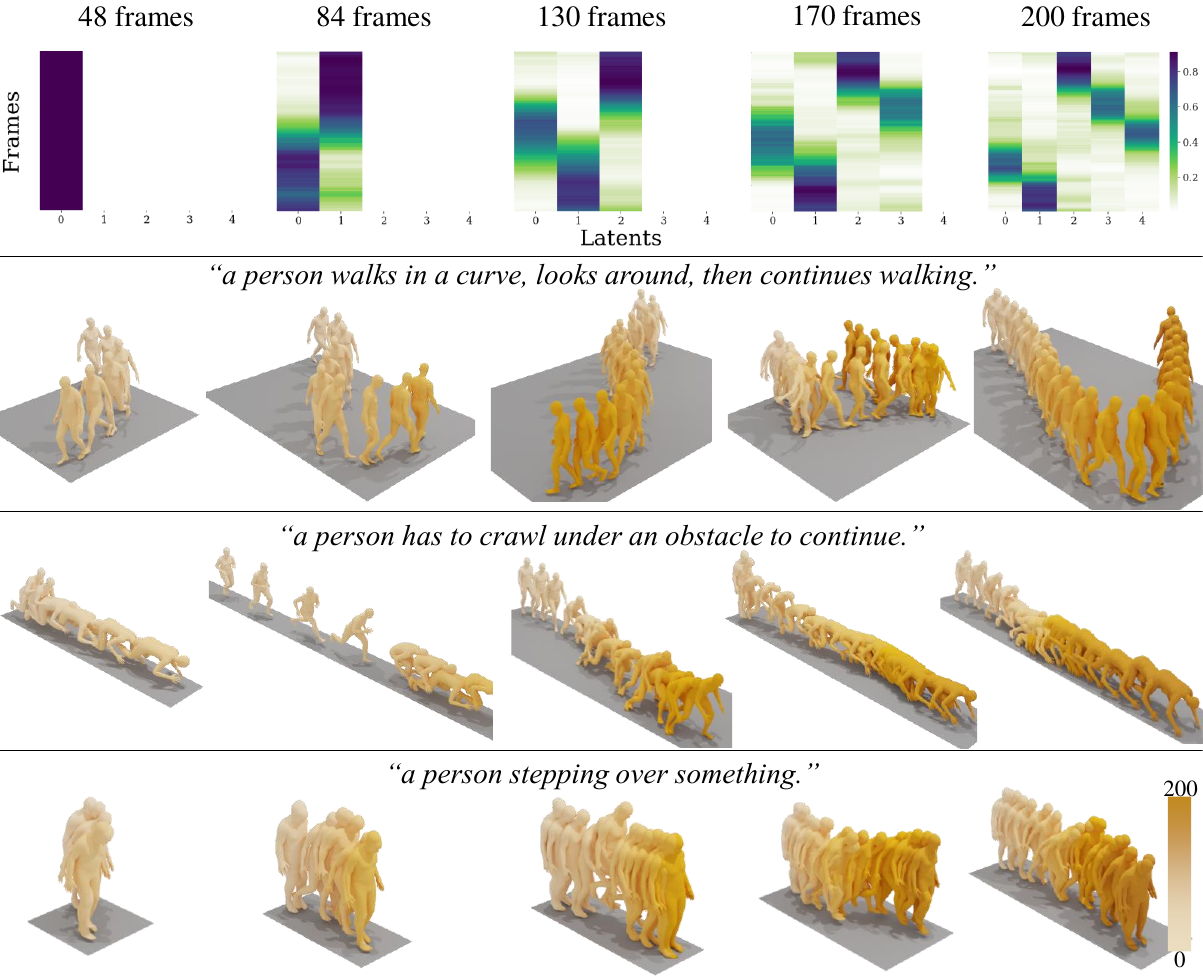}
    \caption{(\textit{Rows}) Generation of motions with the same textual input and varying queried target lengths, alongside (\textit{Top}) the corresponding decoder transformer attention maps where $y-$axis represents the target motion length. Darker colors indicate higher attention scores for each subspace of the latent vector, represented in chunks along the $x-$axis. See Sec \ref{sec:ablation} for the discussion.}
    \label{fig:la_difference}
\end{figure}

\myparagraph{Variable target lengths}. In Fig. \ref{fig:la_difference}, we challenge our model to generate motions for the same textual input but with varying target lengths. At the top, the figure shows the attention maps of the decoder transformer, attending from 1 latent, in the case of querying a 48-frame motion, to 5 latents, for generating motions of 200 frames. Note how, starting from the case of 2 latents, each latent takes responsibility for different contiguous frames in the generated motion. Qualitatively, querying longer target lengths allows \textit{LADiff} to arrange for richer and more careful generated motions.

\begin{table}[!tp]
\caption{Ablation study examining the number of frames per latent $r$, the incorporation of DVAE with a specific noise percentage (\% Noised) both at the input and latent level and the implementation of the Length-Aware framework (LA).}
\label{tab:ablation_la}
\resizebox{1.\textwidth}{!}{
\begin{tabular}{lccccccc}
\hline
\multicolumn{1}{c}{Methods}       &        LA       & $r$                       & Noise (\%)                                 & R Prec. Top 3  $\uparrow$          & FID    $\downarrow$                      & MM-Dist    $\downarrow$                       & Diversity    $\rightarrow$                                 \\ \hline

\multicolumn{1}{l|}{{Real}}  & --- & ---                                      & \multicolumn{1}{c|}{---}                  & {0.797}$^{\pm.002}$             & {0.002}$^{\pm.000}$    & {2.974}$^{\pm.008}$             & {9.503}$^{\pm.065}$                           \\
\multicolumn{1}{l|}{\textit{SOTA}}     & ---                    & ---                                      & \multicolumn{1}{c|}{---}                  & 0.786$^{\pm.002}$                        & 0.112$^{\pm.006}$                        & 3.038$^{\pm.007}$                        & 9.528$^{\pm.071}$                                                \\ \hline

\multicolumn{1}{l|}{}          & \ding{51}                   & 16                                       & \multicolumn{1}{c|}{}                     & {0.778}$^{\pm.001}$       & 0.250$^{\pm.007}$                & 3.129$^{\pm.008}$                & 9.620$^{\pm.083}$                    \\
\multicolumn{1}{l|}{}        & \ding{51}                     & 32                                       & \multicolumn{1}{c|}{}                     & \underline{0.784}$^{\pm.001}$      & \textbf{0.110$^{\pm.004}$}       & \underline{3.077}$^{\pm.010}$       & 9.622$^{\pm.071}$                   \\
\multicolumn{1}{l|}{}       & \ding{51}                      & 48                                       & \multicolumn{1}{c|}{}                     & \textbf{0.792$^{\pm.002}$}       & {0.182}$^{\pm.004}$      & \underline{3.054}$^{\pm.008}$       & {9.795}$^{\pm.076}$         \\
\multicolumn{1}{l|}{}      & \ding{51}                       & 64                                       & \multicolumn{1}{c|}{}                     & {0.780}$^{\pm.002}$       & 0.238$^{\pm.007}$                & 3.114$^{\pm.009}$                & 9.760$^{\pm.093}$            \\
\multicolumn{1}{l|}{\multirow{-5}{*}{Ours}}  & \ding{51}     & all                                      & \multicolumn{1}{c|}{\multirow{-5}{*}{33}} & 0.741$^{\pm.001}$       & 0.445$^{\pm0.10}$                 & 3.375$^{\pm.009}$       & \underline{9.556}$^{\pm.080}$         \\ \hline

\multicolumn{1}{l|}{}        & \ding{51}                     &                                          & \multicolumn{1}{c|}{0}                    & \textbf{0.788$^{\pm.002}$}       & {0.222}$^{\pm.005}$       & \underline{3.062}$^{\pm.009}$      & 9.739$^{\pm.095}$                    \\
\multicolumn{1}{c|}{}         & \ding{51}                    &                                          & \multicolumn{1}{c|}{33}                   & \textbf{0.792$^{\pm.002}$}       & {0.182}$^{\pm.004}$      & \underline{3.054}$^{\pm.008}$       & {9.795}$^{\pm.076}$          \\
\multicolumn{1}{l|}{\multirow{-3}{*}{Ours}}    & \ding{51}   & \multirow{-3}{*}{48}                     & \multicolumn{1}{c|}{50}                   & \textbf{0.787}$^{\pm.002}$                & {0.196}$^{\pm.006}$                & \underline{3.081}$^{\pm.008}$                & 9.764$^{\pm.087}$                           \\ \hline

\multicolumn{1}{l|}{Ours$_{\text{DVAE latent}}$}   &   \ding{51}   &  &        \multicolumn{1}{c|}{}            & 0.735$^{\pm.002}$                & {0.207}$^{\pm.004}$       & 3.389$^{\pm.012}$                & \textbf{9.497}$^{\pm.070}$                    \\
\multicolumn{1}{l|}{Ours$_{\text{DVAE input}}$}             &          \ding{51}       & \multicolumn{1}{c}{\multirow{-2}{*}{48}}                     & \multicolumn{1}{c|}{\multirow{-2}{*}{33}}                  & \textbf{0.792$^{\pm.002}$}       & {0.182}$^{\pm.004}$      & \underline{3.054}$^{\pm.008}$       & {9.795}$^{\pm.076}$              \\ \hline

\multicolumn{1}{l|}{Ours \textit{w/o} \text{LA}}   &     &  &        \multicolumn{1}{c|}{}            & 0.776$^{\pm.001}$                & {0.179}$^{\pm.005}$       & 3.137$^{\pm.011}$                & 9.848$^{\pm.064}$                 \\
\multicolumn{1}{l|}{Ours}             &          \ding{51}       & \multicolumn{1}{c}{\multirow{-2}{*}{48}}                     & \multicolumn{1}{c|}{\multirow{-2}{*}{33}}                  & \textbf{0.792$^{\pm.002}$}       & {0.182}$^{\pm.004}$      & \underline{3.054}$^{\pm.008}$       & {9.795}$^{\pm.076}$          \\ \hline
\end{tabular}
}
\vspace{-.4cm}
\end{table}

\subsection{Ablation Study}
\label{sec:ablation}

Table~\ref{tab:ablation_la} displays ablation studies for the main components of our model, including the number of frames represented by each latent subspace $r$, which determines the activation rate $k$, the use of DVAE (\% Noised), and the adoption of the Length-Aware framework (LA). \\
\myparagraph{Activation rate.} The rate influences the maximum number $K$ of available latent subspaces. Values of $48$ for $r$ and $5$ for $K$ yield the best results and strike a good balance between precision and diversity. \\
\myparagraph{Amount of noise in the DVAE.} Noise is a trade-off between metrics evaluating the movement's precision and the generation process's diversity. The perturbation in DVAE acts on the input sequence, but we explored its application directly on the latent space. The latter approach has a higher impact on the output since the ``filtering'' was entirely charged to the decoder. \\
\myparagraph{Length-awareness} Finally, we showcase the positive impact of our length-aware framework on the results. By comparing it with a model that encodes motion using a fixed dimensionality $K \times D$ (Ours \textit{w/o LA}), we observe an improvement of 1.6\% in R-Precision Top 3 and 2.6\% in Matching Score. In other words, constraining the model to use a variable latent vector dimensionality for varying target lengths improves most of its capability to generate motions closest to the real ones.

\section{In-Depth Analysis}
\label{sec:discussion}
This section investigates \textit{LADiff's} latent space, showcase the activation of length-dependent latent subspaces, and assess the efficiency of the proposed technique.\\

\begin{figure}[!htp] 
    \centering
    \subfloat[Length-aware latent space.]{%
        \includegraphics[width=0.38\textwidth]{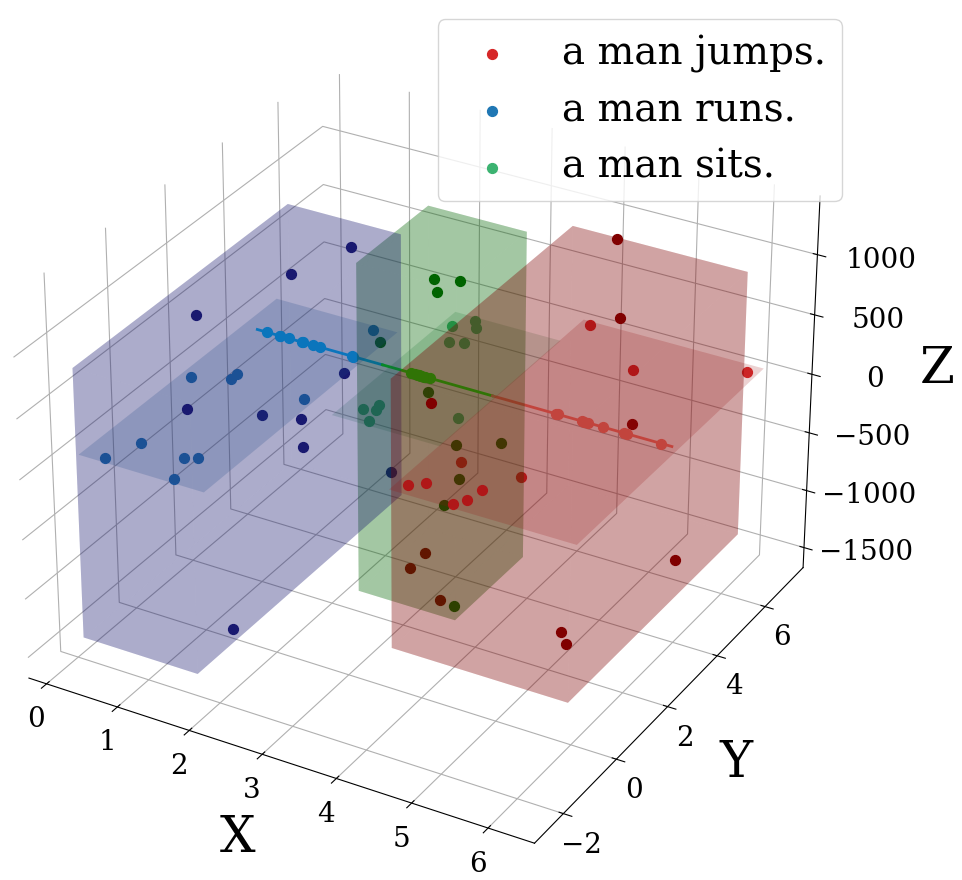}%
        \label{fig:lat_spaces}%
        }%
    \hfill%
    \subfloat[Time/performances trade-off.]{%
        \includegraphics[width=0.52\textwidth]{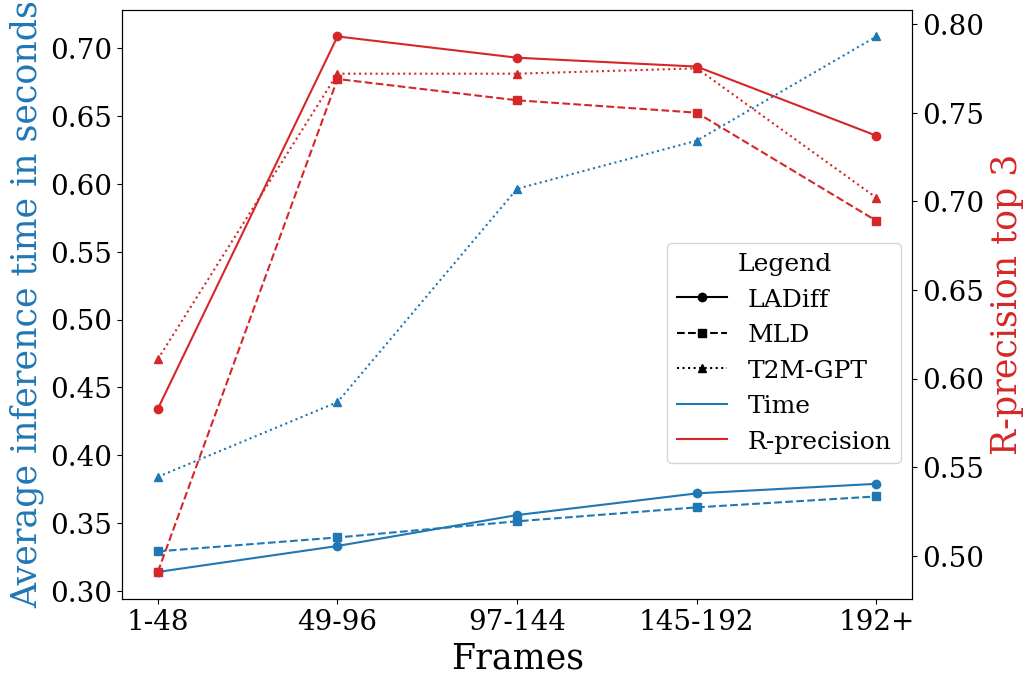}%
        \label{fig:bin}%
        }%
    \caption{Length aware in-depth analysis. Zoom in for details.}
\end{figure}

\myparagraph{Latent Representation.} Fig. \ref{fig:lat_spaces} illustrates the 3D t-SNE reduced dimensionality manifold of the latent space. Each dimension corresponds to a subspace. We generate ten sequences of 30, 96, and 144 frames per action. To estimate the X coordinate, we take the elements of $\mathbf{z}$ associated with the first subspace from each motion and use t-SNE to extract one dimension. Then, we extract from $\mathbf{z}$ the elements associated with the first two subspaces and pad with zeros those without the second to compute with t-SNE the Y component. We use a similar approach to extract the Z component. Short actions generated using a single latent organize themself along a straight line (X-axis). ``Medium'' actions unlock the second latent/dimension with $Y\not = 0$, allowing a displacement of the points along the Y-axis. Finally, the longest motions are free to position in the three-dimensional space, with their orthogonal projection residing in the previous subspace but being free to roam along the third dimension. \textit{LADiff} organizes actions in the latent space by separating the action types along the first subspace and then by clustering the lengths on the available subspaces.\\

\begin{figure}[!ht] 
    \centering
        \includegraphics[width=.95\textwidth]{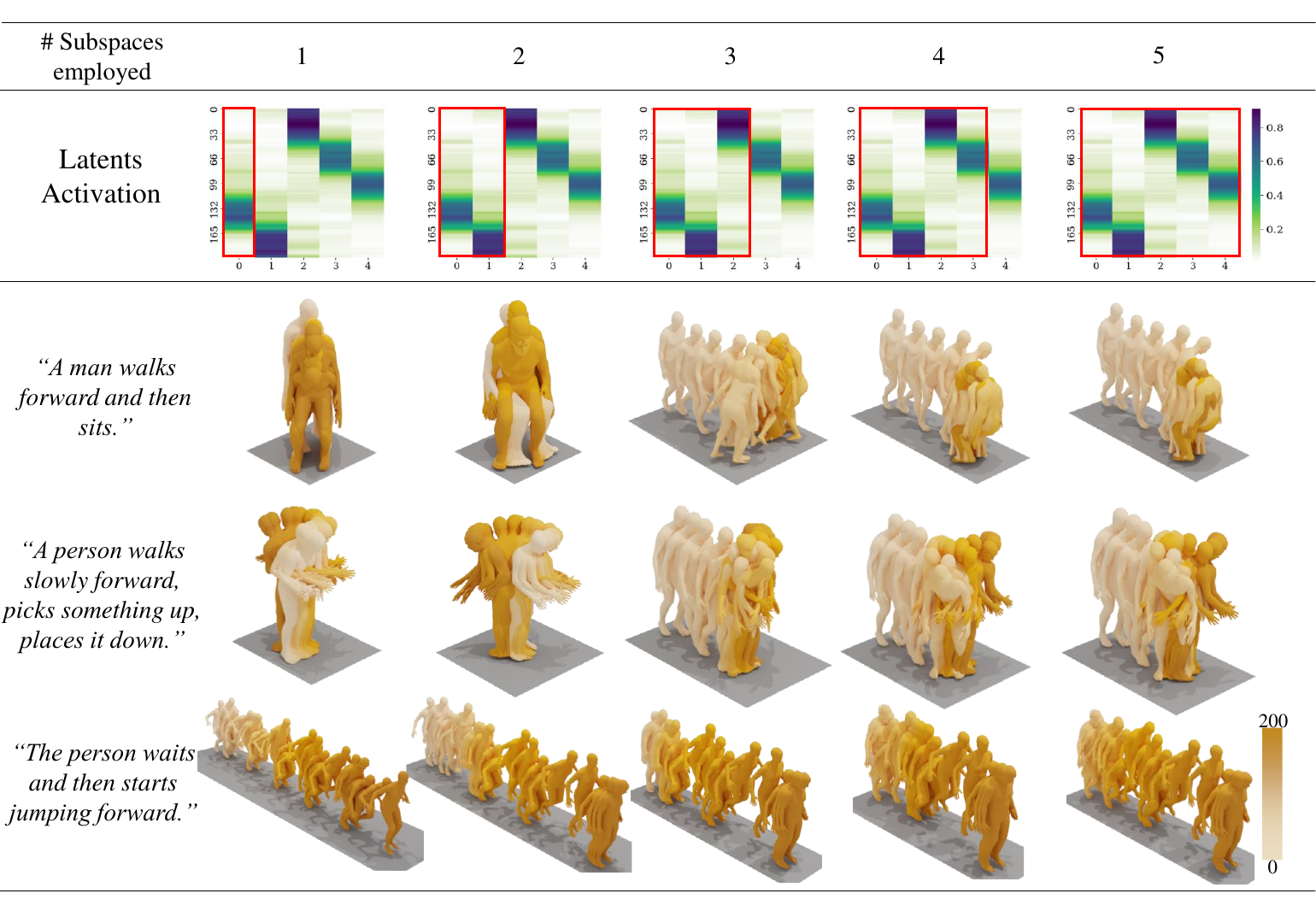}
    \caption{The first row shows the decoder's attention map on the length-aware, denoised latent vectors. Then we depict the generated motion obtained using only the activated latent subspaces selected in red. See Sec~\ref{sec:discussion} for the detailed description.}
    \label{fig:att_maps}
\end{figure}

\myparagraph{Subspace specialization.} Fig. \ref{fig:att_maps} illustrates what happens when we activate only a subset of the subspaces to generate a motion that is long and that requires the entire latent space. From left to right, we activate the subspaces one by one. Each latent vector is accountable for almost precisely $r$ contiguous frames. We observe a gradual transition between these blocks, possibly to connect different actions and movements and guarantee a smooth generation. When only the first or two latents are used (left-most two examples), the generated motion is focused on the action of \textit{sitting}  in the second row. The third latent subspace introduces the \textit{walking} part; the motion is complete but has substantial jitter and unnatural movements. The fourth and fifth latent subspaces fill the gap between the initial and final movements, generating smooth transition and realistic behavior. \\
 The two text conditionings \textit{A person walks slowly forward, picks something up, places it down}, depicted in the third row, and \textit{The person waits and then starts jumping forward} in the fourth row corroborate our analysis as we observe the first two subspaces attending to the final part of the action (e.g., \textit{``picks something up, places it down''} or \textit{``jumping forward''}). The third contributes to generating the initial frames. The fourth and fifth fill the gaps, harmonizing these two parts.\\
 The different responsibility of subspaces for different lengths ensures that longer sequences add new styles and dynamics.

\myparagraph{Performance and Inference time across lengths.}
In Figure~\ref{fig:bin}, we illustrate our model's influence on inference time and R-Precision Top 3 across different generated motion lengths (5-bins) of the HumanML3D dataset. We compare against the models T2M-GPT~\cite{zhang2023t2mgpt} and  MLD~\cite{chen2023mld}, top performers for which the code is available. Compared to T2M-GPT, \textit{LADiff} outperforms precision scores (red curves) in every generated motion length (up to 4\%) except for the shortest sequences where T2M-GPT prevails by a short margin. However, \textit{LADiff} is considerably faster (blue curves) in every setting. Compared to MLD, \textit{LADiff} has superior precision performance (up to 10\%) with comparable inference times.\\

\myparagraph{Length-aware motion dynamics.}
We question the correspondence between the length of the generated motions and the statistics of the motion dynamics, specifically considering velocity and acceleration of the body joints. In Table~\ref{tab:dynamic}, we report these statistics for all the qualitatives presented in the main paper and the supplementary material. For MLD, the averages of acceleration and velocity remain unchanged as the generated length varies. By contrast, for \textit{LADiff} the statistics increase by 44.4\% and 17.3\%, respectively, when the motion is shorter. We confirm therefore that \textit{LADiff} produces different motion styles for different lengths. In Table~\ref{tab:actions_dynamic}, we repeat the analysis on the same analysis on a controlled set of atomic actions (\textit{``A man sits''}, \textit{``A man walks''}, and \textit{``A man throws''}). Each action has peculiar velocities and acceleration, but the relative changes with motions are similar.\\

\myparagraph{Limitations.}
\textit{LADiff} assumes a linear correlation between motion lengths and the number of exploited subspaces. While functional and performant, we advocate for future research on this aspect.
\noindent\begin{minipage}[!t]{.5\textwidth}%
\centering
\captionof{table}{Comparison of dynamics on\\ input text.}
\label{tab:dynamic}
\resizebox{0.92\textwidth}{!}{
\begin{tabular}{@{}l|cc|cc|cc|@{}}
\cline{2-7}
                                           & \multicolumn{2}{c|}{Avg. Vel. ($m/s$)}     & \multicolumn{2}{c|}{Avg. Acc. ($m/s^2$)}    & \multicolumn{2}{c|}{Max Acc. ($m/s^2$)}     \\ \cline{2-7} 
                                             & 84   & \multicolumn{1}{c|}{170}   & 84   & \multicolumn{1}{c|}{170}   & 84   & \multicolumn{1}{c|}{170}   \\ \hline
 \multicolumn{1}{|c|}{MLD}    &   0.39   & \multicolumn{1}{c|}{0.39}     &   0.06    & \multicolumn{1}{c|}{0.05}     &  0.31  & \multicolumn{1}{c|}{0.33}        \\
 \multicolumn{1}{|c|}{\textit{LADiff}} &   0.61   & \multicolumn{1}{c|}{0.52}      & 0.13 & \multicolumn{1}{c|}{0.09}     &    0.61    & \multicolumn{1}{c|}{0.55}         \\ \hline
\end{tabular}}
\hfill%
\end{minipage}%
\begin{minipage}[!t]{.5\textwidth}%
\captionof{table}{Comparison of dynamics on atomic actions with \textit{LADiff}.}
\hfill%
\label{tab:actions_dynamic}
\resizebox{0.92\textwidth}{!}{
\begin{tabular}{@{}c|ccc|ccc|ccc|@{}}
\cline{2-10}
                              & \multicolumn{3}{c|}{Avg. Vel. ($m/s$)}       & \multicolumn{3}{c|}{Avg. Acc. ($m/s^2$)}  &
                              \multicolumn{3}{c|}{Max Acc. ($m/s^2$)}
                              \\ \hline
\multicolumn{1}{|c|}{Actions} & 48 & 84 & \multicolumn{1}{c|}{170} &  48 & 84 & \multicolumn{1}{c|}{170}   &  48 & 84 & \multicolumn{1}{c|}{170}  \\ \hline
\multicolumn{1}{|c|}{Sit}     & 0.27   & 0.17   & \multicolumn{1}{c|}{0.10}      &  0.05  &  0.04  & \multicolumn{1}{c|}{0.02}    &  0.16  &  0.16  & \multicolumn{1}{c|}{0.14}     \\
\multicolumn{1}{|c|}{Walk}    & 1.31   & 1.01   & \multicolumn{1}{c|}{0.72}        & 0.11   &  0.09  & \multicolumn{1}{c|}{0.06}   & 0.19   &  0.18  & \multicolumn{1}{c|}{0.14}      \\
\multicolumn{1}{|c|}{Throw}   &  0.16  &  0.15  & \multicolumn{1}{c|}{0.13}        &  0.04  & 0.04   & \multicolumn{1}{c|}{0.03}   &  0.14  & 0.15   & \multicolumn{1}{c|}{0.13}     \\ \hline
\multicolumn{1}{|c|}{Mean}    &  0.58    & 0.44   &          \multicolumn{1}{c|}{0.31}    &    0.06    & 0.05   &          \multicolumn{1}{c|}{0.03}     &    0.16   & 0.16   &          \multicolumn{1}{c|}{0.14}                     \\ \hline
\end{tabular}}
\end{minipage}%

\section{Conclusion}
We have presented Length-Aware Latent Diffusion (\textit{LADiff}), a novel approach for generating motion sequences guided by textual descriptions with dynamics and style conditioned by the target motion length. To do so, \textit{LADiff} leverages a customized latent space with subspaces representing sequences of different lengths. Addressing length awareness may be a first step in coding control mechanisms for the generated sequence attributes. \\

\myparagraph{Acknowledgements.} We acknowledge financial support from
ItalAI S.r.l., the PNRR MUR project PE0000013-FAIR and from the Sapienza grant RG123188B3EF6A80 (CENTS). 


%
%
\bibliographystyle{splncs04}
\bibliography{main}
\end{document}